\documentclass[letterpaper, 10 pt, conference]{ieeeconf}  % Comment this line out if you need a4paper

\IEEEoverridecommandlockouts                              % This command is only needed if 
\usepackage[linesnumbered,ruled,vlined, noend]{algorithm2e}
\usepackage{mathtools}
\usepackage{amsmath,tabularx}   % \langle-- for \eqref
\usepackage{subcaption}
\usepackage{bm}
\usepackage[T1]{fontenc}
\usepackage{graphicx}
\usepackage{booktabs}
\usepackage{cite}
\usepackage{amssymb}
\usepackage{xcolor}
\usepackage{soul}
\usepackage{float}
\usepackage{hyperref}
\usepackage[capitalize]{cleveref}
\usepackage{tikz} 
\usetikzlibrary{positioning}
\usetikzlibrary {shapes.geometric}
\usetikzlibrary{fit}
\usepackage{newfloat}
\usepackage{listings}
\usepackage{multirow}
\let\labelindent\relax
\usepackage{enumitem}
\usepackage{xcolor}

\newcommand{\Rho}{\mathrm{P}}
\newcommand{\Real}{\mathbb R}
\newcommand{\Natural}{\mathbb N}
\newcommand{\Powerset}{\mathcal P}

\newcommand{\X}{\mathcal{X}}
\newcommand{\Y}{\mathcal{Y}}

\newcommand{\T}{\mathcal{T}}

\newcommand{\odAIRL}{\textsf{oDec-AIRL}}
\newcommand{\odMDP}{\textsf{oDec-MDP}}
\newcommand{\odPPO}{\textsf{oDec-PPO}}
\newcommand{\dAIRL}{Dec-AIRL}

\title{\LARGE \bf
Open Human-Robot Collaboration 
using Decentralized Inverse Reinforcement Learning 
}

\author{Prasanth Sengadu Suresh$^{1,2}$, Siddarth Jain$^{1}$, Prashant Doshi$^{2}$, Diego Romeres$^{1}$% \langle-this % stops a space
\thanks{$^{1}$Mitsubishi Electric Research Laboratories (MERL), Cambridge, MA, USA. 
        {\tt\small \{sjain,romeres\}@merl.com }}%
\thanks{$^{2}$School of Computing, University of Georgia, Athens, GA, USA. This research was completed during P. Suresh’s internship at MERL.{\tt\small \{ ps32611,pdoshi\}@uga.edu }}
}

\begin{document}

\maketitle
\thispagestyle{empty}
\pagestyle{empty}

%%%%%%%%%%%%%%%%%%%%%%%%%%%%%%%%%%%%%%%%%%%%%%%%%%%%%%%%%%%%%%%%%%%%%%%%%%%%%%%%
\begin{abstract}

The growing interest in human-robot collaboration (HRC), where humans and robots cooperate towards shared goals, has seen significant advancements over the past decade. While previous research has addressed various challenges, several key issues remain unresolved. Many domains within HRC involve activities that do not necessarily require human presence throughout the entire task. Existing literature typically models HRC as a \emph{closed} system, where all agents are present for the entire duration of the task. In contrast, an \emph{open} model offers flexibility by allowing an agent to enter and exit the collaboration as needed, enabling them to concurrently manage other tasks. In this paper, we introduce a novel multiagent framework called \odMDP{}, designed specifically to model open HRC scenarios where agents can join or leave tasks flexibly during execution. We generalize a recent multiagent inverse reinforcement learning method - \dAIRL{} to learn from \emph{open} systems modeled using the \odMDP{}. Our method is validated through experiments conducted in both a simplified toy firefighting domain and a realistic dyadic human-robot collaborative assembly. 
Results show that our framework and learning method improves upon its closed system counterpart.

\end{abstract}

%%%%%%%%%%%%%%%%%%%%%%%%%%%%%%%%%%%%%%%%%%%%%%%%%%%%%%%%%%%%%%%%%%%%%%%%%%%%%%%%
\section{INTRODUCTION}
\label{sec:Introduction}

As the landscape of artificial intelligence and robotics continues to evolve, the collaboration between humans and robots has gained significant importance. This partnership harnesses the unique and often complementary strengths of both humans and robots~\cite{villani2018survey}. Robots, equipped with sensory perception and intelligent decision-making abilities, play a crucial role as collaborators in enhancing efficiency, precision, and creativity across diverse fields. The Human-Robot Collaboration (HRC) paradigm is not geared towards replacing human labor; rather, it focuses on augmenting human capabilities, empowering individuals to tackle challenges that were once considered insurmountable. Previous studies in HRC have explored human and robotic agent interactions using multiagent decision-making frameworks~\cite{dahiya2023survey}. However, many existing multiagent models typically operate within a \emph{closed system} framework, where a fixed group of human and robotic agents collaborate from initiation to completion of a task. This closed system approach lacks the flexibility provided by an \emph{open} system, where agents can dynamically join or leave the task at various stages as needed. This characteristic of openness is referred to as \emph{agent openness}~\cite{eck2023decision}.

In this work, we consider domains where only a subset of tasks necessitate human collaboration. Recognizing human limitations in time and energy, an \emph{open} system permits humans to seamlessly join and collaborate with robots when their input is essential. Such a framework is termed as an \emph{open-HRC system (OHRCS)}. Despite its effectiveness, OHRCS introduces several challenges. For instance, in collaborative table assembly tasks with numerous components (\cref{fig:furniture-description}), there can exist multiple valid sequences for assembly, with only a small subset of tasks requiring human intervention. Subsequently, there may be a particular order that minimizes the time and effort of the human. In such situations, the primary difficulty lies in crafting a model that is sufficiently sophisticated to encompass a wide range of potential behaviors. This multiagent model needs to accurately represent the behaviors of both the existing team of agents and any new agents that join, as well as the nature of the task itself. Furthermore, given that many real-world scenarios are decentralized~\cite{goldman2003decentralized}---meaning each agent may lack complete information about the others--the model must account for these system dynamics. The second major challenge involves reward engineering. The reward function must be intricate enough to encourage behaviors that effectively solve the task optimally, while also balancing the increased cost incurred by utilizing human assistance. Such reward shaping is non-trivial. 

In this paper, we describe our inverse reinforcement learning (IRL) based approach to the aforementioned OHRCS challenges and make two key contributions. First, we present \odMDP{}, a novel multiagent decision-making framework to model agent openness in OHRCS. Second, we develop a novel IRL technique - \odAIRL{} that generalizes a recent decentralized IRL method - \dAIRL{}~\cite{sengadu2023dec}, to learn the underlying reward function and its corresponding vector of policies using the \odMDP{} as the behavioral model. We validate our contributions on two domains: a simulated Urban Firefighting~\cite{eck2023decision,kakarlapudi2022decision,chandrasekaran2016individual, eck2020scalable} scenario and a realistic physical human-robot collaborative furniture assembly.

%%%%%%%%%%%%%%%%%%%%%%%%%%%%%%%%%%%%%%%%%%%%%%%%%%%%%%%%%%%%%%%%%%%%%%%%%%%%%%%%

\section{BACKGROUND}
\label{sec:background}

Multiagent IRL typically models the expert using multiagent generalizations of the Markov decision process (MDP) such as multiagent MDP (MMDP), Markov game, or decentralized MDP (Dec-MDP). Due to the nature of HRC being decentralized and collaborative, a Dec-MDP is appropriate to model the collaborative expert team. A two-agent Dec-MDP can be formally defined as a tuple $$\mathcal{DM} \triangleq \langle S, A, T, R \rangle$$ 
where the global state, $S = S_i \times S_j$. Here, $S_i$ and $S_j$ are the locally observed states of the two agents $i$ and $j$, which when combined yield the complete global state of the system; $A = A_i \times A_j$ is the set of joint actions of the two agents; $T: S \times A \times S \rightarrow [0,1]$ is the transition function of the multi-agent system; and $R: S \times A \rightarrow \mathbb{R}$ is the common reward function~\footnote{This Dec-MDP describes a locally fully observable model whose local states when combined yield the fully observable global state~\cite{Goldman03}.}.  In IRL, the latter is unknown, whereas the rest of the elements are usually known. As such, the agents know their local state and any common task attributes; agents act independently while optimizing a task-centric reward function~\cite{melo2011decentralized}. Let $\X^E$ be the set of expert demonstrations and a complete trajectory $X^E \in \X^E$ is given by, $X^E = (\langle s_i^0,s_j^0\rangle, \langle a_i^0,a_j^0\rangle, \langle s_i^1,s_j^1\rangle, \langle a_i^1,a_j^1\rangle, ..., \langle s_i^\T,s_j^\T\rangle, \langle a_i^\T,a_j^\T\rangle).$

%------------------------------------------------------------------------------------------------------------------------------------
\subsection{Review of Decentralized Adversarial IRL}
\label{subsec:dec-airl}
%------------------------------------------------------------------------------------------------------------------------------------

Decentralized adversarial IRL (\dAIRL{})~\cite{sengadu2023dec} generalizes the single-agent deep-IRL method - adversarial IRL~\cite{fu2018learning} (that works on the principle of maximum causal entropy) to learn a common reward function for the team, from expert demonstrations. AIRL uses a discriminator $D_{\bm{\theta}} (X)$ to learn a function $f_{\bm{\theta}} (X)$~\cite{fu2018learning} which at convergence approximates the advantage function corresponding to the expert's policy. \dAIRL{} analytically represents the discriminator as: $D_{\bm{\theta}}(X) = \frac{e^{f_{\bm{\theta}}(X)}}{e^{f_{\bm{\theta}}(X)} + \pi(X)}$ and the reward update rule is given as
\begin{align}
R_{\bm{\theta}}(X) \leftarrow \log D_{\bm{\theta}}(X) - \log(1 - D_{\bm{\theta}}(X)).
\label{eqn:airl-reward-update}
\end{align}

\cref{eqn:airl-reward-update} when simplified yields: $f_{\bm{\theta}}-\log(\pi)$, which is the entropy-regularized reward formulation. In the underlying Dec-MDP~\cite{goldman2004decentralized}, each agent only has access to their local state and some general task attributes. Dec-PPO - a decentralized generalization of the popular RL method - Proximal Policy Optimization~\cite{schulman2017proximal}, is used as \dAIRL{}'s forward-rollout technique. Dec-PPO uses the centralized training, decentralized execution paradigm where the centralized critic network updates its value function as a squared-error loss:
\begin{align*}
    &L^{VF}_{t}(\omega) = (V^{\pi_\omega}(s^{t}) - \Hat{V}_{t}^{targ} )^2.
\end{align*}
where $\Hat{V}_{t}^{targ}$ is the per-episode discounted reward-to-go and $V^{\pi_{\omega}}(s^{t})$ is the predicted value of global state $s^{t}$ and $\omega$ is the policy weight vector. For a dyadic system with agents $i$ and $j$, the policy loss of agent $i$ is given by
\begin{small}
\begin{align*}
&L^{CLIP}_i(\omega) = \mathop{\mathbb{E}}_{\pi_{\omega, i}} \Bigl[min\Bigl(\lambda_i ~ A^{\pi_{\omega}}, clip(\lambda_i,1-\epsilon, 1+\epsilon)~A^{\pi_{\omega}}\Bigr)\Bigr] 
\end{align*}
\end{small}
where $\lambda_i^t = \frac{\pi_{\omega,i}(a_{i}^t|s_{i}^t)}{\pi_{\omega,i}^{old}(a_{i}^t|s_{i}^t)}$ is the importance sampling ratio. $L^{CLIP}(\omega)$ provides a pessimistic bound over the final objective by using a surrogate objective that picks the minimum of the clipped and unclipped objectives. By clipping the importance sampling ratio, the incentive of moving $\lambda^{t}$ outside the interval $[1-\epsilon, 1+\epsilon]$ is reduced. $A_{\pi_{\omega}}^t$ is calculated with respect to the reward estimates $R_{\bm{\theta}}(X)$ from \cref{eqn:airl-reward-update}. This clipped surrogate objective, combined with the policy entropy, handles the explore-exploit dilemma. The policy entropy loss is given as:
\begin{align*} 
L^{ENT}_{i}(\omega) =  \sigma H [\pi_{\omega,i}(s_{i}^{t}) ]. 
\end{align*}
where $H$ is the policy entropy and $\sigma$ is the entropy hyperparameter. The total loss is then given as:
\begin{align} 
L_{i}(\omega) =  L^{CLIP}_i(\omega) + L^{ENT}_{i}(\omega).
\label{eqn:Dec-PPO-loss}
\end{align}
The policy loss, entropy loss, and total loss apply analogously for agent $j$. At the end of training, the discriminator and generator return the learned common reward function and the converged vector of policies respectively.

%%%%%%%%%%%%%%%%%%%%%%%%%%%%%%%%%%%%%%%%%%%%%%%%%%%%%%%%%%%%%%%%%%%%%%%%%%%%%
\section{Open Human-Robot Collaboration}
\label{sec:method}

In this section, we introduce the novel \odMDP{} and describe how \odAIRL{} uses it to model OHRC problems.
%------------------------------------------------------------------------------------------------------------------------------------
\subsection{Open Collaboration Model}
\label{subsec:interaction-model}
%------------------------------------------------------------------------------------------------------------------------------------
\begin{figure}
\centering
\begin{tikzpicture}[node distance={15mm}, thick, main/.style = {draw, circle}]
\node[main] (1) {$c^t$}; 
\node[main] (2) [below left of = 1] {${\bm s}^{t}_{c^t}$}; 
\node[main] (3) [right of = 2] {${\bm a}^{t}_{c^t}$}; 
\node[main, scale=0.9] (4) [above right of = 3] {$c^{t+1}$}; 
\node[main, scale=0.9] (5) [right=0.75cm of 3] {${\bm s}^{t+1}_{c^{t+1}}$}; 
\draw[->] (1) -- (2); 
\draw[->] (1) -- (3); 
\draw[->] (2) -- (3); 
\draw[->] (1) -- (4); 
\draw[->] (2) to [out=300, in=225, looseness=0.5] (5); 
\draw[->] (3) -- (4); 
\draw[->] (3) -- (5); 
\draw[->] (4) -- (5); 
\end{tikzpicture} 
\caption{\small The \odMDP{} graphical model for two timesteps $t$ and $t+1$. Given the collab team ID $c^t$ at timestep $t$, ${\bm s}^{t}_{c^t}$ is formed by combining the local states of all agents in $c^t$. All agents' local actions from $c^t$ combined form ${\bm a}^{t}_{c^t}$, which leads to $c^{t+1}$, given $c^t$. $c^{t+1}$, ${\bm a}^{t}_{c^t}$ and ${\bm s}^{t}_{c^t}$ together lead to the next state ${\bm s}^{t+1}_{c^{t+1}}$ at time $t+1$.}
\label{fig:PGM}
\vspace{-0.25cm}
\end{figure}
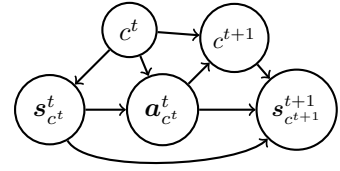

\odMDP{} generalizes Dec-MDP to model agent openness in a decentralized, collaborative setting. Formally, 
%our \odMDP{} is defined as:
\begin{align}
\odMDP{} \triangleq \langle Ag, C, \bm{S}, \bm{A}, \Gamma, T, R, \rho \rangle \nonumber
\end{align}
%where     
\begin{itemize}[leftmargin=*, itemsep=0in, topsep=0in]
    \item $Ag$ is the finite set of all agents and $|Ag| = N$ is the maximum number of agents;
    \item  $C : \Powerset(Ag) \rightarrow  \Natural$ assigns a unique number identifier to each collaborating team and $\Powerset$ denotes the powerset excluding the empty set. For convenience, we let $C$ also denote the set of all assigned identifiers;
    \item Global state space $\bm{S} = \bigcup_{c=1}^{c=|C|} S_{c}$ where $c \in C$ and $S_c$ denotes the set of states of the team identified by $c$;
    \item Global action space $\bm{A} = \bigcup_{c=1}^{c=|C|} A_{c}$ where $c \in C$ and $A_c$ denotes the set of {\em joint} actions of the team identified by $c$. For instance, if team $c$ involves agents $i$ and $j$ whose action sets are $A_i$ and $A_j$, respectively, then $A_c = A_i \times A_j$; 
    \item Team transition model $\Gamma: C \times \bm{A} \times C \rightarrow [0,1]$ gives the distribution of the new teams given the current team and action letting agent(s) enter or exit;
    \item $T = \{T_c,T'_c ~|~ c = 1, 2, \ldots, |C|\}$, where {\em intra-team} state transition model $T_c: S_c \times A_c \times S_c \rightarrow [0,1]$ gives the distribution over the team's next state and {\em inter-team} state transition model $T'_c: S_c \times C' \times S_{c'} \rightarrow [0,1]$ gives the distribution over the next team's state. Both are available for all $c, c' \in C$;
    \item Common reward function shared by all agents in each team $c$, $R_c \triangleq R(S_c, A_c, c)$ and $R_c : S_c \times A_c \rightarrow \Real$;     
    \item Start state and team prior distribution $\rho: S \times C \rightarrow [0,1]$.
\end{itemize}

An open teamwork  trajectory of length $\T$ contains the collaborating team ID, team state, and team action at each time step: 
\begin{small}    
\begin{align*}
    &X^E \triangleq \left( \langle  c, {\bm s}_c, {\bm a}_c \rangle^1, \langle c, {\bm s}_c, {\bm a}_c \rangle^2, \langle c', {\bm s}_{c'}, {\bm a}_{c'} \rangle^3
    % \right. \nonumber \\ \left.
    \ldots \langle c'', {\bm s}_{c''}, {\bm a}_c'' \rangle^{\T} \right). \nonumber
\end{align*}
\end{small}

Notice from the trajectory that the starting team with ID $c$ persists for the first two timesteps followed by a change to team $c'$. If the team with ID $c$ at time step $t=1$ is a dyad with agents $i$ and $j$, then the policy $\bm{\pi}$, the team state ${\bm s}^{t}_c$, and team action ${\bm a}^{t}_c$ are vectors of the two individual agents' policies, their partial states, and their actions respectively:
\begin{align}
    {\bm \pi}_c \triangleq \langle \pi_i, \pi_j\rangle;~
    {\bm s}^{t}_c \triangleq \langle s^{t}_{i}, s^{t}_{j} \rangle;~ \mbox{and }
    {\bm a}^{t}_c \triangleq \langle a^{t}_{i}, a^{t}_{j} \rangle. \nonumber
\end{align}

The likelihood of the first two time steps of $X^E$ is obtained using the parameters of \odMDP{} as:
\begin{small}
\begin{align}
&\Pr(c, {\bm s}^1_c, {\bm a}^1_c, c, {\bm s}^2_{c}, {\bm a}^2_c) \nonumber \\
&= \Pr(c, {\bm s}^2_c, {\bm a}^2_c | c, {\bm s}^1_c, {\bm a}^1_c) \Pr({\bm a}^1_c | c, {\bm s}^1_c) \Pr(c, {\bm s}^1_c)\nonumber\\
&= \Pr({\bm a}^2_c | c, {\bm s}^2_c) \Pr(c | c, {\bm a}^1_c) \Pr({\bm s}^2_c |{\bm s}^1_c, {\bm a}^1_c) \Pr({\bm a}^1_c | c, {\bm s}^1_c) \Pr(c, {\bm s}) \nonumber \\
&= \Pr(a^2_i | c, s^2_i) \Pr(a^2_j | c, s^2_j)~\Gamma(c, {\bm a}^1_c, c)~T_c({\bm s}^1_c, {\bm a}^1_c, {\bm s}^2_c) \nonumber\\
~&\quad \times \Pr(a^1_i | c, s^1_i) \Pr(a^1_j | c, s^1_j)\Pr(c, {\bm s}^1_c)\nonumber\\ 
&= \underbrace{\pi_i(a^2_i | c, s^2_i)}_{\text{Policy of i at t=2}}~\underbrace{\pi_j(a^2_j | c, s^2_j)}_{\text{Policy of j at t=2}}~\underbrace{\Gamma(c, {\bm a}^1_c, c)}_{\text{Team transition}}~\underbrace{T_c({\bm s}^1_c, {\bm a}^1_c, {\bm s}^2_c)}_{\text{intra-team state transition}}\nonumber\\
&\quad \times~\underbrace{\pi_i(a^1_i | c, s^1_i)}_{\text{Policy of i at t=1}}~\underbrace{\pi_j(a^1_j | c, s^1_j)}_{\text{Policy of j at t=1}}~\underbrace{\rho(c, {\bm s}^1_c)}_{\text{given prior}}. \label{eqn:likelihood-1}
\end{align}
\end{small}
\noindent A locally fully observable Dec-MDP lets each agent's policy condition its action on the agent's partial view of the state. Next, we show how the likelihood is obtained for the second and third timesteps of $X^E$ when the team changes. Its derivation proceeds analogously to the above steps:
\begin{small}
\begin{align}
&\Pr(c, {\bm s}^2_c, {\bm a}^2_c, c', {\bm s}^3_{c'}, {\bm a}^3_{c'}) = \pi_i(a^3_i | c', s^3_i)~\pi_j(a^3_j | c', s^3_j)~\Gamma(c, {\bm a}^2_c, c')\nonumber\\
&\quad \times~T'_c({\bm s}^2_c, c', {\bm s}^3_{c'})~\pi_i(a^2_i | c, s^2_i)~\pi_j(a^2_j | c, s^2_j)~\rho(c, {\bm s}^2_c).
\label{eqn:likelihood-2}
\end{align}
\end{small}
\noindent The key difference between Eqs.~\ref{eqn:likelihood-1} and~\ref{eqn:likelihood-2} is that the latter involves the inter-team transition function $T'_c$ due to the change of team from time step $t=2$ to $t=3$. 
For the sake of completeness, we also show below how the value of a given team state at timestep $t$, ${\bm s}_c^t$, is obtained, which defines the value function of our \odMDP{}:
\begin{small}    
\begin{align*}
&V({\bm s}^t_c)
= \max_{{\bm a}^t_c} ~\mathbb{E}_{c',{\bm s}^{t+1}_{c'}} \left[ R({\bm s}^t_c, {\bm a}^t_c, c) + \gamma  V({\bm s}^{t+1}_{c'})|{\bm s}^t_c, c \right]\\
&= \max_{{\bm a}^t_c} ~R({\bm s}^t_c, {\bm a}^t_c, c) + \gamma \sum_{C'} \sum_{{\bm S}_{c'}} \Pr(c', {\bm s}^{t+1}_{c'}|c, {\bm s}^t_c, {\bm a}^t_c)~V({\bm s}^{t+1}_{c'})\\
&= \max_{{\bm a}^t_c} ~R_c + \gamma \sum_{C'} \sum_{{\bm S}_{c'}}~\Gamma(c, {\bm a}^t_c, c')~T_c'({\bm s}^t_c, c', {\bm s}^{t+1}_{c'})~V({\bm s}^{t+1}_{c'}). \nonumber
\end{align*} 
\end{small}

%------------------------------------------------------------------------------------------------------------------------------------
\subsection{Method}
\label{subsec:open-dec-airl}
%------------------------------------------------------------------------------------------------------------------------------------

The discriminator $D_{\bm \theta}(X)$ of \odAIRL{} learns a common reward function $R_c$ contingent on  $c, {\bm s}, \text{and } {\bm a}$. This common reward function is then used by \odPPO{} to learn a vector of policies (one for each agent). \odAIRL{} minimizes the reverse KL divergence between the learner's and expert's marginal teamID-state-action distribution $KL(\Rho_\pi(c, {\bm s}, {\bm a})~||~\Rho_{exp}$ $(c, {\bm s}, {\bm a}))$. The \odAIRL{} algorithm shown in \cref{alg:main} takes the \odMDP{} ($\mathcal{ODM}$) without the reward and transition functions, and the expert trajectories $\X^E$, as input. The goal is to learn a common reward function $R_c$ for the task based on $\X^E$, and the corresponding vector of learned policies.

%%%%%%%%%%%%%%%%%%%%%%%%%%%%%%%%%%%%%%%%%%%%%%%%%%%%%%%%%%%%%%%%%%%%%%%%%%%%%%%%%%%
\vspace{0.02cm}
\begin{algorithm}[ht!]
\begin{small}
\caption{\odAIRL}
\label{alg:main}
\SetKwInput{KwInput}{Input}
\SetKwInput{KwOutput}{Output}
\DontPrintSemicolon

\KwInput{$\mathcal{ODM}$ sans $R_c$, $\Gamma$ and $T$; Exp trajs $\X^E$.}
\KwOutput{Learned common reward function $R_c$.}

Initialize decentralized policy ${\bm \pi}_c$, Discriminator $D_{\bm{\theta}}$.\;

\For{$iter \leftarrow 0$ \KwTo $train$\textunderscore $iters$}{
    Generate joint trajectories $\hat{\X}$ using ${\bm \pi}_c$.\;
    Sample joint $\langle c, {\bm s}, {\bm a}\rangle$ minibatches $\hat{\Y}$ and $\Y^{E}$ from $\hat{\X}$ and $\X^E$, respectively.\;

    \For{$ep\gets0$ \KwTo $discriminator$\textunderscore $epochs$}{
        Train discriminator $D_{\bm{\theta}}$ to minimize $KL(\Rho_\pi(c, {\bm s}, {\bm a})~||~\Rho_{exp}(c, {\bm s}, {\bm a}))$.\;
    }

    Update reward $R_c \leftarrow \log D_{\bm{\theta}}(X) - \log (1 - D_{\bm{\theta}}(X))$.\;

    \For{$ep\gets0$ \KwTo $generator$\textunderscore $epochs$}{
        Train generator $G(R_c) \leftarrow$ \odPPO{}.\;
    }

    Get updated policy ${\bm \pi}_c \leftarrow G(R_c)$.\;
}

\Return{$R_c$, ${\bm \pi}_c$.}
\end{small}
\end{algorithm}

The algorithm begins by initializing a random decentralized policy vector ${\bm \pi}_c$ (line 1), and a discriminator $D_{\bm{\theta}}$ with random weights $\bm{\theta}$. Learning continues until the end of training iterations (line 2-10). Each iteration, it generates joint trajectories $\hat{\X}$ using the current policy vector ${\bm \pi}_c$. It then samples minibatches of $\langle c,{\bm s},{\bm a}\rangle$ from $\hat{\X}$ and $\X^E$ to yield $\hat{\Y}$ and $\Y^E$ respectively (line 4). It trains $D_{\bm{\theta}}$ using $\hat{\Y}$ and $\Y^E$ to minimize the reverse KL divergence between the expert and learned distributions (line 6). Using $D_{\bm{\theta}}$'s confusion it extracts an updated reward $R_c$ (line 7). $R_c$ is then provided as an input to the generator $G(R_c)$ using \odPPO{} which learns the forward rollout vector of policies (line 10). Finally, the learned reward function $R_c$ and policy vector ${\bm \pi}_c$ are returned (line 11).

%%%%%%%%%%%%%%%%%%%%%%%%%%%%%%%%%%%%%%%%%%%%%%%%%%%%%%%%%%%%%%%%%%%%%%%%%%%%%

\begin{figure*}[tbh!]
    \begin{subfigure}{0.164\textwidth}
      \centering
      \includegraphics[width=\linewidth]{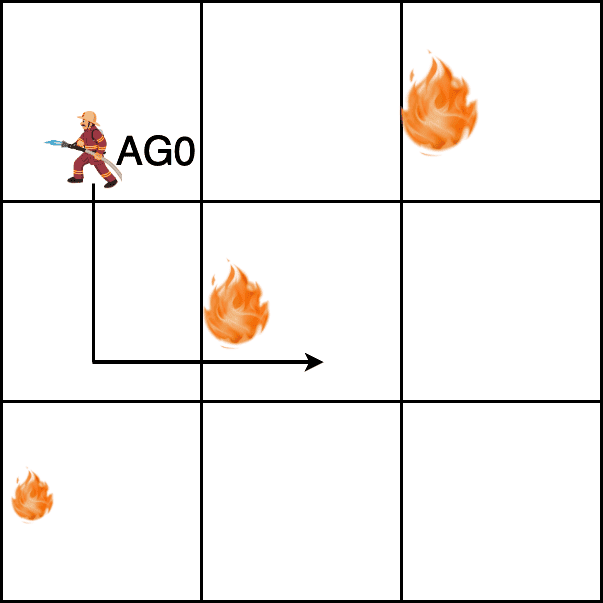}
      \caption{}
      \label{fig:UFF_ag0_travels}
    \end{subfigure}\hfill
    \begin{subfigure}{0.164\textwidth}
      \centering
      \includegraphics[width=\linewidth]{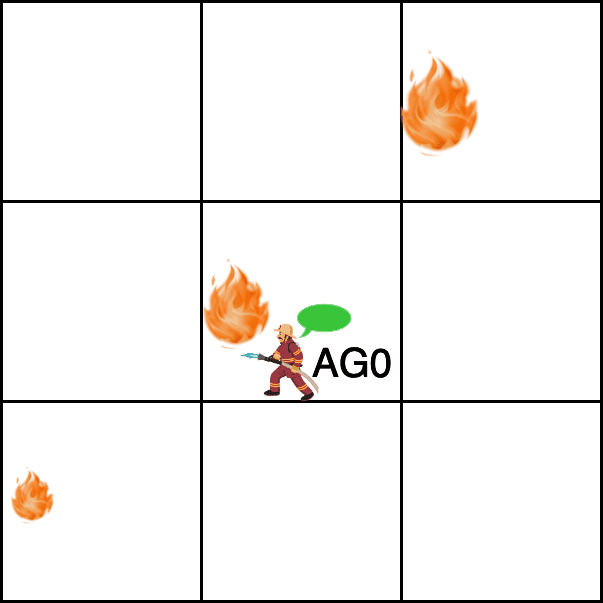}
      \caption{}
      \label{fig:UFF_ag0_calls_ag1}
    \end{subfigure}\hfill
    \begin{subfigure}{0.164\textwidth}
      \centering
      \includegraphics[width=\linewidth]{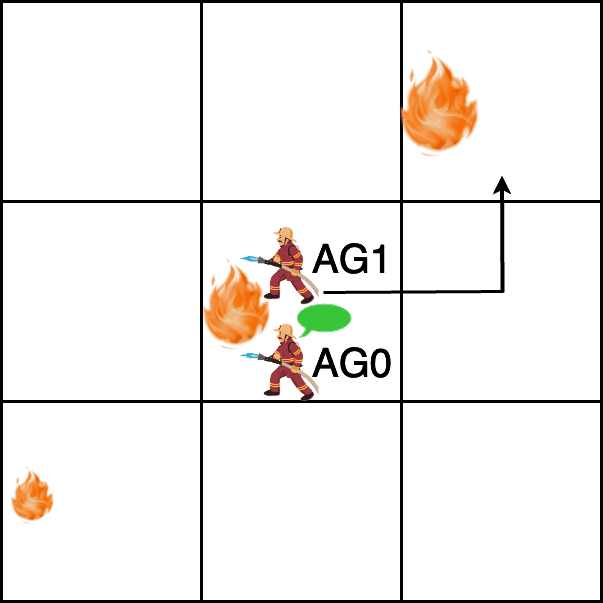}
      \caption{}
      \label{fig:UFF_ag0_calls_ag2_ag1_travels}
    \end{subfigure}    
    \begin{subfigure}{0.164\textwidth}
      \centering
      \includegraphics[width=\linewidth]{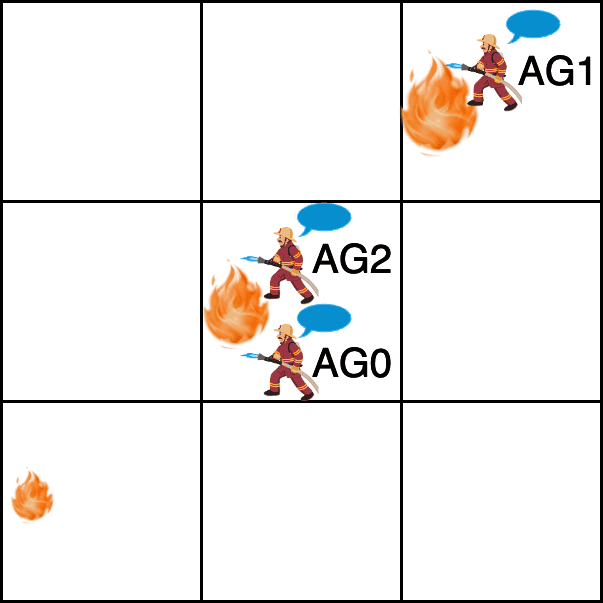}
      \caption{}
      \label{fig:UFF_ag0_ag2_ag1_extinguish}
    \end{subfigure}\hfill
    \begin{subfigure}{0.164\textwidth}
      \centering
      \includegraphics[width=\linewidth]{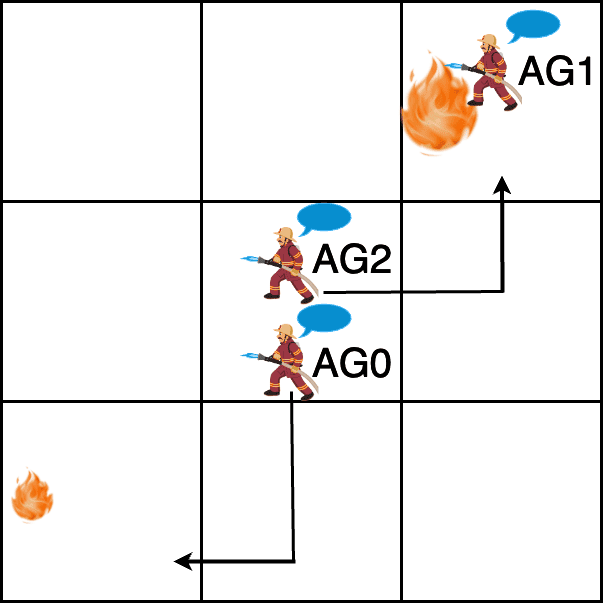}
      \caption{}
      \label{fig:UFF_ag0_ag2_move_ag1_extinguish}
    \end{subfigure}\hfill
    \begin{subfigure}{0.164\textwidth}
      \centering
      \includegraphics[width=\linewidth]{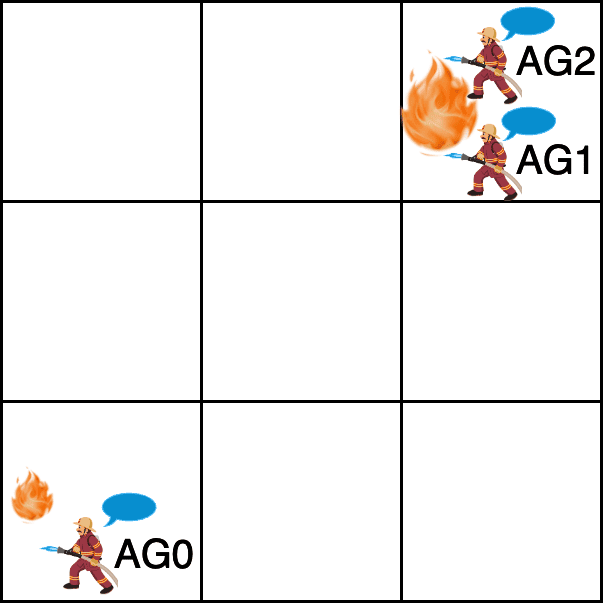}
      \caption{}
      \label{fig:UFF_ag0_extinguish_ag2_ag1_extinguish}
    \end{subfigure}\hfill
    \caption{\small A complete episode of the Urban Firefighting domain with learned \odAIRL{} policies. Colored bubbles green and blue represent CallAgent and Extinguish actions respectively. In Figure \ref{fig:UFF_ag0_travels}, Agent 0 heads towards the medium fire, and Figure \ref{fig:UFF_ag0_calls_ag1}, calls Agent 1 to join. Subsequently, Agent 1 moves to the large fire in Figure \ref{fig:UFF_ag0_calls_ag2_ag1_travels}, while Agent 0 calls Agent 2. In Figure \ref{fig:UFF_ag0_ag2_ag1_extinguish}, all agents extinguish fires at their locations. After extinguishing the medium fire, Agent 0 moves to the small fire, and Agent 2 assists Agent 1 with the large fire in Figure \ref{fig:UFF_ag0_ag2_move_ag1_extinguish}. Figure \ref{fig:UFF_ag0_extinguish_ag2_ag1_extinguish} shows agents at their final locations performing the Extinguish action. This visualization was built using PyGame~\cite{mcgugan2007beginning}.}
    \label{fig:urban-firefighting}
\vspace{-0.2cm}
\end{figure*}

\begin{table*}[tbh!]
\caption{Urban Firefighting learned policies comparison}
\label{table:urban-firefighting}
\setlength{\tabcolsep}{8pt} % Adjust the column spacing as needed
\centering
\begin{small}
\begin{tabular}{|c|c|c|c|c|c|} 
\hline
\multicolumn{6}{|c|}{Average of $10000$ timesteps} \\
\hline
\multirow{2}{*}{Max Num Agents} & Method & \multicolumn{3}{c|}{\small Num Steps Active Per Eps} & {\small Episode Reward} \\
\cline{2-6}
& & \small Agent0 & \small Agent1 & \small Agent2 &  \\
\hline
\hline
\multirow{2}{*}{2} & \textbf{\odAIRL{}} & {\small \boldmath $18.06$ $\pm$ $1.13$} & {\small \boldmath $13.06$ $\pm$ $0.61$} & - & {\small \boldmath $32.22$ $\pm$ $0.59$} \\
\cline{2-6}
 & \dAIRL{} & {\small $16.87$ $\pm$  $0.09$} & {\small $16.87$ $\pm$  $0.09$} & - & $30.37$ $\pm$ $0.07$ \\
 \hline
 \multirow{2}{*}{3} & \textbf{\odAIRL{}} & {\small \boldmath $12.36$ $\pm$ $0.57$} & {\small \boldmath $9.27$ $\pm$ $0.12$} & {\small \boldmath $8.27$ $\pm$ $0.02$} & {\small \boldmath $37.86$ $\pm$ $0.13$} \\
\cline{2-6}
 & \dAIRL{} & {\small $10.33$ $\pm$ $1.29$} & {\small $10.33$ $\pm$ $1.29$} & {\small $10.33$ $\pm$ $1.29$} & $35.73$ $\pm$ $0.70$ \\
\hline
\end{tabular}
\end{small}
\vspace{-0.25cm}
\end{table*}

\section{Experiments}
In this section, we consider two domains, a popular simulated toy domain - Urban Firefighting~\cite{eck2023decision,kakarlapudi2022decision,chandrasekaran2016individual, eck2020scalable}, and a realistic dyadic collaborative human-robot furniture assembly domain. In both domains, we focus on the agent being called in at the right time. Nonetheless, our method applies analogously to agents exiting the task. To establish the efficacy of an open system over a closed one, we compare the learned behavior of \odAIRL{} with an ablation study using \dAIRL{}, on both domains.
%------------------------------------------------------------------------------------------------------------------------------------
\subsection{Urban Firefighting}
%------------------------------------------------------------------------------------------------------------------------------------

In this domain, the goal is for the agent(s) to extinguish fires within a $3$x$3$ grid as efficiently as possible. There are three active fires: a large, medium, and small fire as shown in \cref{fig:urban-firefighting} with intensities $0.9$, $0.6$, and $0.3$ respectively at the beginning. Each firefighter has 4 cardinal direction actions, a CallAgent action, and an Extinguish action. A CallAgent action calls one agent at a time (up to $3$ agents) and the Extinguish action at a fire location reduces its intensity by $0.1$. Each agent's local state contains their location, the fire locations and intensities, and the number of teammates at their location. Agents need to decide whether to call another agent to extinguish the fires effectively. The challenge in this domain is that the learned reward function must maintain a delicate balance between the cost incurred by having additional agents and the reward of extinguishing fires sooner. 

\noindent \textbf{Results:} To provide expert data for training, we generated simulated trajectories based on human preferences, of $10^6$ total timesteps for both \odAIRL{} and \dAIRL{}, trained both methods for $10^7$ timesteps, and compared their best-learned behaviors using average episode reward and the average number of timesteps each agent is active per episode. As can be observed from \cref{table:urban-firefighting}, in the behavior learned by \dAIRL{}, all agents are present throughout the task duration as expected, and in turn incur a higher step-cost. In comparison, \odAIRL{} policies only engage the second and third agents for $9.27$ and $8.27$ steps, respectively, in the three-agent case and accrue a higher average episodic reward. Note that the total number of steps per episode is slightly greater in an open system compared to a closed one since agents are being called in one at a time as needed.
%------------------------------------------------------------------------------------------------------------------------------------
\subsection{Collaborative Furniture Assembly}
%------------------------------------------------------------------------------------------------------------------------------------

\begin{figure*}
    \centering
    \includegraphics[width=.95\linewidth]{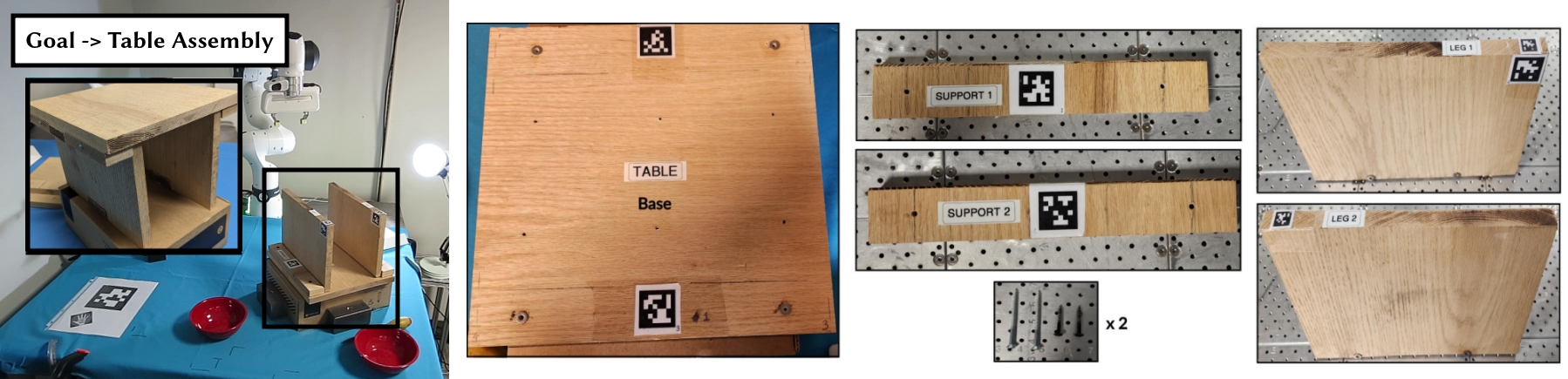}
    \caption{\small A collaborative table assembly task that involves placing and screwing together various wooden components. \textit{Left: } The assembled table. \textit{Right: } The individual components required for assembly, including the table base, two supports, two legs, and the necessary screws.}
    \label{fig:furniture-description}
    \vspace{-0.2cm}
\end{figure*}

\begin{figure*}
    \centering
      \includegraphics[width=0.95\linewidth]{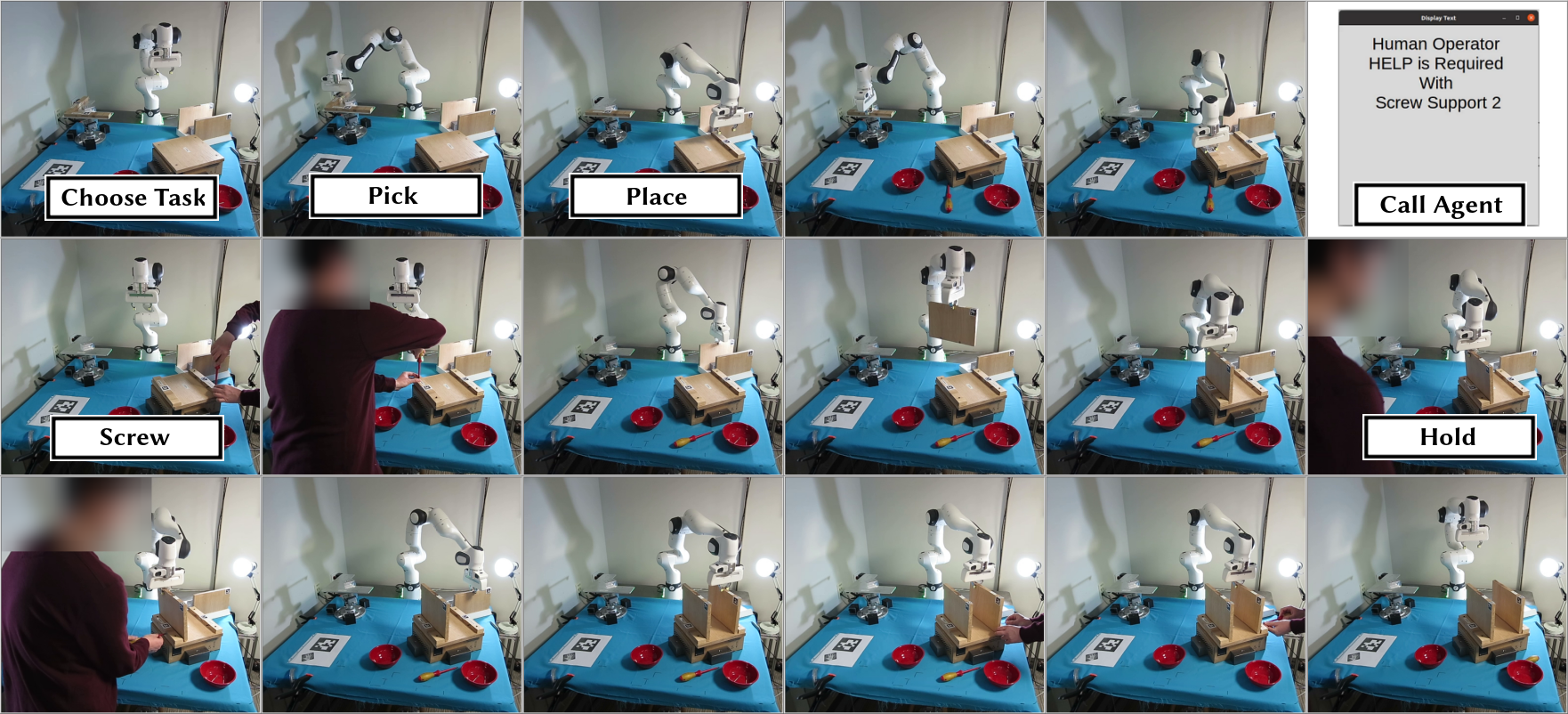}
    \caption{\small Snapshots capturing key moments from a typical trial in the human-robot pilot study. At each step, the robotic agent executes actions based on its learned policy \odAIRL{}, while the human participant is encouraged to emulate the learned human policy. Examples of each action are illustrated upon their initial occurrence. Following the completion of the `Place' action for Support1, the robot prompts for human assistance through a pop-up notification for the `Call Agent' action.}
    \label{fig:hrc-assembly-task2}
    \vspace{-0.2cm}
\end{figure*}

In this realistic dyadic HRC furniture assembly task, the goal is to assemble a table consisting of multiple parts: base, leg-support$1$, leg-support$2$, leg$1$, leg$2$, two screws for each leg-support to connect them to the base, and two screws for each leg to screw them into their respective supports, as shown in \cref{fig:furniture-description}. The task can be completed in multiple ways. One may position both leg-supports on the base and screw them in before positioning their corresponding legs and screwing the legs into their respective supports. Alternatively, one may position leg-support$1$, screw it into the base, place the leg$1$, screw it into the leg-support$1$, and analogously repeat the sequence for the other parts to complete the assembly. Notice that the positioning actions can be done independently by the robot, while the screwing action requires the assistance of a human. While the speed of assembly could be increased by having the human position parts in parallel from the beginning, the step-cost incurred due to the human's presence would be quite high. This means that the learned reward function must optimize completing the assembly sooner and the step-cost due to the human's presence. In other words, the optimal behavior must only call the human into the task when imperative.

\begin{table*}[tbh!]
\caption{\textbf{HRC Furniture Assembly learned policies comparison}}
\label{table:hrc-assembly}
\setlength{\tabcolsep}{8pt} % Adjust the column spacing as needed
\centering
\begin{small}
\begin{tabular}{|c|c|c|c|c|} 
\hline
\multicolumn{5}{|c|}{Average of $10000$ timesteps} \\
\hline
\multirow{2}{*}{Max Num Agents} & Method & \multicolumn{2}{c|}{\small Num Steps Active Per Eps} & {\small Episode Reward} \\
\cline{2-5}
& & \small Robot & \small Human &  \\
\hline
\hline
\multirow{2}{*}{2} & \textbf{\odAIRL{}} & {\small \boldmath $19.61 \pm 0.94$} & {\small \boldmath $13.20 \pm 0.98$} & {\small \boldmath $4.06 \pm 0.58$} \\
\cline{2-5}
 & \dAIRL{} & {\small $16.34 \pm 0.97$} & {\small $16.34 \pm 0.97$} & $1.74 \pm 0.60$ \\
\hline
\end{tabular}
\end{small}
\vspace{-0.25cm}
\end{table*}

Each agent has 8 discrete actions: \emph{ChooseTask} - This randomly assigns a valid next task to perform, \emph{Pick} - Agent picks up the current part, \emph{Place} - Agent places the current part at the goal location, \emph{HoldInPlace} - Agent holds the current part steadily at its current location, \emph{ScrewIn} - Agent screws the current part into place, \emph{CallAgent} - This calls the human into the task, \emph{ResetTask} - Agent places the current part back to its original location, \emph{NoOp} - No action. The local state of each agent in the expert's \odMDP{} consists of three discrete variables: \emph{TaskName} - which takes a valid task name from eleven discrete values when a ChooseTask action is performed; \emph{TaskStatus} - which describes the current status of the task through one of seven discrete values; \emph{Collab} - which provides the current collaboration level between unavailable, partial and full collaboration. This domain was developed as a text-based discrete-state-action Gym environment on MA-Gym~\cite{magym} based on domain knowledge of the assembly task.
\begin{figure*}[tbh!]
    \centering
    \includegraphics[width=1.0\linewidth]{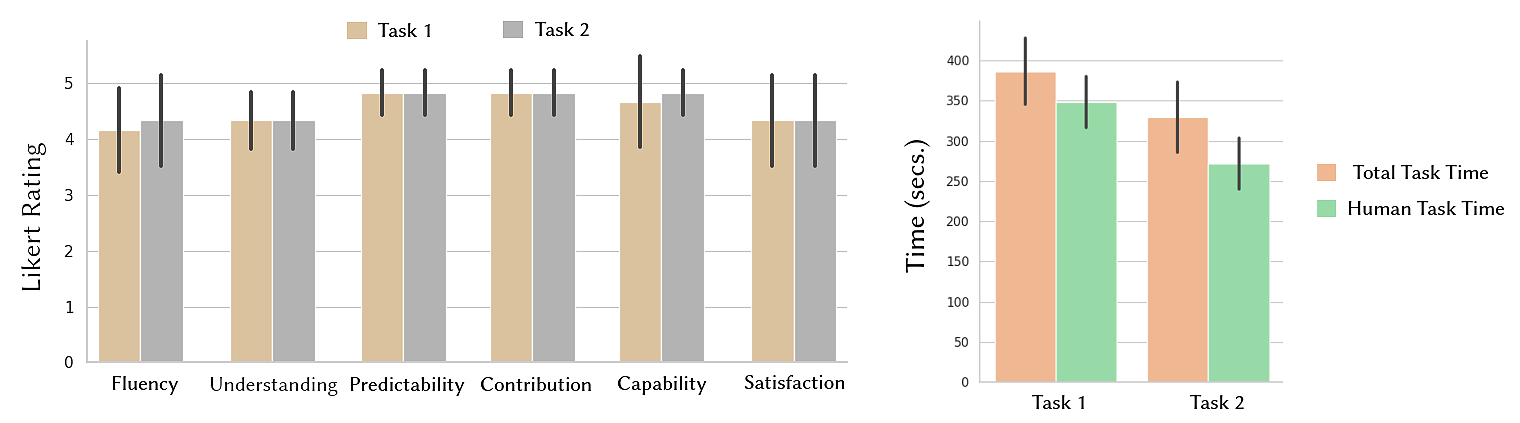}
    \caption{\small \textit{Left: }Findings for subjective measures on a 5-point scale. \textit{Right: }The average total duration of tasks and the average time allocated to human agents starting from the Call Agent action.}
    \label{fig:HRC-results}
    \vspace{-0.25cm}
\end{figure*}
After completing a screwing subtask, the human can either stay idle (doing \emph{NoOp}) until the robot needs help again or participate by positioning other parts in parallel. The upside to the latter is that the task is completed sooner and the team receives a better reward; the downside is that if the human is engaged in a different task while the robot requires help, the human must perform a \emph{ResetTask} action before helping the robot, extending the task duration. 

\noindent \textbf{Simulation Experiments:} We generated simulated expert trajectories of $10^6$ total timesteps and trained both \odAIRL{} and \dAIRL{} for $10^7$ timesteps. These simulated trajectories were generated based on human preferences and domain knowledge to generate team behavior that optimally completes the task while minimizing human time and effort. In these trajectories, the human agent on average intervenes 4 times per episode to perform the screwing action. When the human is called in depends on the task order, which is randomly selected in each episode to maximize policy exploration. We compared their best-learned behaviors using average episode reward and the average number of timesteps the human and robot are present per episode. Notice that by following the \dAIRL{} policies, the human is active for the entire task duration, while with \odAIRL{} policies, the human is only active for $13.2$ timesteps per episode as shown in Table \ref{table:hrc-assembly}. Naturally, \odAIRL{} policies accrue a better episode reward than the baseline policies.
  
\begin{table}[tbh!]
    \centering
    \caption{Subjective Measures}
    \label{table:hrc-questionnaire}
    \begin{tabular}{>{\raggedright\arraybackslash}p{0.45\textwidth}}
        \toprule
        \midrule
        \textbf{Fluency} \\
        The robot and I collaborated fluently to accomplish the task.\\
        \midrule
        \textbf{Understanding} \\
        I feel the robot had a good understanding of the task. \\
        \midrule
        \textbf{Predictability} \\
        I was never surprised by the robot's actions. \\
        \midrule
        \textbf{Contribution} \\
        The robot called for my assistance at the appropriate time.  \\
        \midrule
        \textbf{Capability} \\
        I feel that my time and effort were valued by the robot. \\
        \midrule
        \textbf{Satisfaction} \\
        I feel satisfied with the performance of the system. \\
        \midrule
        \bottomrule        
    \end{tabular}
    \vspace{-0.2cm}
\end{table}

\noindent \textbf{Real-world HRC experiment:} In our pilot study, we conducted experiments where human subjects collaborated in assembling a table using a Franka Emika Research-3 7-DoF robotic arm. The robot's control stack was implemented as a finite state machine to manage and monitor its states and actions. Initially trained neural network policies were exported to a CSV file, which was then incorporated into the finite state machine to guide the robot's actions at each state. We recruited 6 participants under a within-subjects design. Participants received instructions on the assembly task, the robot's action capabilities, and its objectives. They were tasked to participate in the assembly and cooperate with the robot, especially when it requested assistance via a `CallAgent' action displayed on an on-screen graphical interface popup. To aid in marker-based state estimation, we deployed an externally mounted Intel Realsense D435 camera with April Tags on the assembly components. Additionally, we integrated human hand tracking using a deep learning model~\cite{zhang2020}, which was meticulously calibrated within the experimental setup. An AprilTag positioned just outside the workspace was used to signal the completion of actions, which participants activated after each task step. 

For manipulated variables, we shuffled task sequences and selected two variations of the table assembly task. Consequently, each participant performed the assembly twice. Task $1$ entailed positioning leg-support$2$ and screwing it into the base, then positioning leg$2$ and securing it into its support, and analogously for leg-support$1$ and leg$1$ to complete the task. Conversely, task $2$ involved positioning leg-supports $2$ and $1$, screwing them into the base, positioning leg$1$, securing it into its support, and repeating the process for leg$2$ to complete the assembly. Our focus was on examining the framework for real-world open human-robot collaboration and determining the optimal timing of the `CallAgent' action based on task variations. For other measures, we measured task completion times and the duration spent by the human across both tasks (\cref{fig:HRC-results}, right). We created six statements for subjective evaluation (\cref{table:hrc-questionnaire}), and asked participants to rate their level of agreement on a 5-point Likert scale (inspired by the Godspeed Series Questionnaire ~\cite{bartneck2008godspeed}).

All trials of the study were successful. \cref{fig:hrc-assembly-task2} shows the Franka Research-$3$ robot assembling the table as per task order $2$, calling the human for assistance at the appropriate time using the on-screen popup, after which the human and the robot collaboratively complete the rest of the assembly. \cref{fig:HRC-results} (left), shows the subjective ratings of the real-world experiments. \cref{fig:HRC-results} (right), shows the quantitative measures of the real-world experiments. Task $1$ takes an average of $386.76 \pm 41.19$ secs for completion, while Task $2$ takes $348.42 \pm 32.28$ secs. Through the `Call Agent' action, on average, human agents only spend $329.49 \pm 43.98$ secs on Task $1$ and $271.82 \pm 31.55$ secs on Task $2$ demonstrating successful OHRC through an average time saving of approximately 18.39\% for the human across both tasks.

%%%%%%%%%%%%%%%%%%%%%%%%%%%%%%%%%%%%%%%%%%%%%%%%%%%%%%%%%%%%%%%%%%%%%%%%%%%%%
\section{Related Work}

This work marks the initial endeavor within HRC to introduce a learn-from-observation approach for contexts involving agent openness (AO). Whereas methods tackling planning, modeling, and learning challenges in open-agent systems exist, we outline their limitations for OHRCS.
%------------------------------------------------------------------------------------------------------------------------------------

Previous methods addressing HRC tasks adopt different modeling approaches. Nikolaidis et al.~\cite{nikolaidis2012human} and Chen et al.~\cite{chen2020trust} use a single-agent POMDP, while Giacomuzzo et al.~\cite{giacomuzzo2024decaf} use a single-agent Discrete-Event MDP to represent the domain. In later work, Nikolaidis et al.~\cite{nikolaidis2017game} employ a two-player Markov game to characterize HR teams, where humans adjust their behavior based on evolving perceptions of the robot's capabilities. Seo and Unhelkar~\cite{seo2022semi} divided scenarios into an agent model and a task model. A centralized task model captures the task attributes while the agent Markov model represents the mental states of the other interacting agents. Wang et al.~\cite{wang2022co} employ an MMDP to simulate a handover and combined human-robot manipulation task, learning joint policies. More recent work~\cite{sengadu2023dec}  utilizes a Dec-MDP to model HRC scenarios, where the human and robot are aware only of their local states. The robot policy is evaluated on a real-world collaborative produce sorting task~\cite{suresh2022marginal}. It is worth noting that these studies assumed closed systems for HRC, which limits their applicability to OHRCS.

More generally, Eck, Doshi and Soh~\cite{eck2023decision} define open agent systems as a multiagent scenario where some elements of the system dynamically change over time. They outline three primary types of openness: agent openness (AO), where agents can join or leave during the task; type openness (TO), involving dynamic changes in the frames of agents present; and task openness (TaO), which allows changes in current tasks. Accurately representing open systems with uncertainties regarding tasks, active agents, and their characteristics is paramount. Framed by these forms of openness, Chen et al.~\cite{chen2015considering} investigates openness from an ad hoc standpoint, concluding through simulations that AO enhances performance whereas TaO complicates learning. Another work ~\cite{chandrasekaran2016individual} employs the IPOMDP-Lite framework~\cite{hoang2013interactive} to simulate AO, validating it in simulated scenarios such as wildfire suppression. A representation close to our \odMDP{} framework is the Open Dec-POMDP~\cite{cohen2017open} to model AO. A key distinction is that they incorporate AO via coalitions, where the coalition transition function depends on the previous coalition and disregards the actions of agents. This assumption may be impractical when coalition transitions depend on the policies of agents. Another distinction is that the Open Dec-POMDP does not model state transitions due to team member's actions and team transitions. 

To study AO in settings involving many agents, Eck et al.~\cite{eck2020scalable} enhances scalability by selectively modeling neighbors and extrapolating behavior from this modeling to others. A CI-POMDP model involving communication has also been utilized~\cite{kakarlapudi2022decision}, incorporating agents' messages into belief updates validated via simulations. More recently, a RL-based decentralized actor-critic method based on the I-POMDP framework learns policies for open-organization challenges where employees may be hired or removed. Rahman et al.~\cite{rahman2023general} proposes a partially observable open stochastic Bayesian game to model AO, employing graph-based policy learning evaluated across simulated domains. These approaches share a common drawback: they typically assess their methodologies using small, simulated toy environments, which may not effectively scale to real-world scenarios. Techniques relying on I-POMDPs or game theory may not be adequate for collaborative teamwork, while those for ad hoc teamwork overlook the dynamic nature of agents entering and exiting tasks. Moreover, the absence of evaluation in physical systems like real robots makes it challenging to gauge their efficacy in operational HRC contexts.

%%%%%%%%%%%%%%%%%%%%%%%%%%%%%%%%%%%%%%%%%%%%%%%%%%%%%%%%%%%%%%%%%%%%%%%%%%%%%
\section{Conclusions and Future Work}

In this paper, we addressed agent openness (AO) in human-robot collaboration systems, and introduced a novel multiagent model, \odMDP{}, and a new IRL technique, \odAIRL{}, which uses expert demonstrations to learn a reward function for solving OHRC problems. Future work may relax the assumption of full observability to handle environmental occlusions or sensor noise and consider a larger set of agents to validate scalability. We also plan to explore type openness (TO) in OHRC, such as how fatigue affects human collaboration levels. OHRCS enhances seamless and efficient collaboration between human and robotic agents, opening numerous future research directions. 

%%%%%%%%%%%%%%%%%%%%%%%%%%%%%%%%%%%%%%%%%%%%%%%%%%%%%%%%%%%%%%%%%%%%%%%%%%%%%%%%

\bibliographystyle{IEEEtran}
\bibliography{iros24}

\begin{thebibliography}{10}
\providecommand{\url}[1]{#1}
\csname url@rmstyle\endcsname
\providecommand{\newblock}{\relax}
\providecommand{\bibinfo}[2]{#2}
\providecommand\BIBentrySTDinterwordspacing{\spaceskip=0pt\relax}
\providecommand\BIBentryALTinterwordstretchfactor{4}
\providecommand\BIBentryALTinterwordspacing{\spaceskip=\fontdimen2\font plus
\BIBentryALTinterwordstretchfactor\fontdimen3\font minus \fontdimen4\font\relax}
\providecommand\BIBforeignlanguage[2]{{%
\expandafter\ifx\csname l@#1\endcsname\relax
\typeout{** WARNING: IEEEtran.bst: No hyphenation pattern has been}%
\typeout{** loaded for the language `#1'. Using the pattern for}%
\typeout{** the default language instead.}%
\else
\language=\csname l@#1\endcsname
\fi
#2}}

\bibitem{villani2018survey}
V.~Villani, F.~Pini, F.~Leali, and C.~Secchi, ``Survey on human--robot collaboration in industrial settings: Safety, intuitive interfaces and applications,'' \emph{Mechatronics}, vol.~55, pp. 248--266, 2018.

\bibitem{dahiya2023survey}
A.~Dahiya, A.~M. Aroyo, K.~Dautenhahn, and S.~L. Smith, ``A survey of multi-agent human--robot interaction systems,'' \emph{Robotics and Autonomous Systems}, vol. 161, p. 104335, 2023.

\bibitem{eck2023decision}
A.~Eck, L.-K. Soh, and P.~Doshi, ``Decision making in open agent systems,'' \emph{AI Magazine}, 2023.

\bibitem{goldman2003decentralized}
C.~V. Goldman and S.~Zilberstein, ``Decentralized control of cooperative agents,'' \emph{JAIR}, 2003.

\bibitem{sengadu2023dec}
P.~Sengadu~Suresh, Y.~Gui, and P.~Doshi, ``Dec-airl: Decentralized adversarial irl for human-robot teaming,'' in \emph{Proceedings of the 2023 International Conference on Autonomous Agents and Multiagent Systems}, 2023, pp. 1116--1124.

\bibitem{kakarlapudi2022decision}
A.~Kakarlapudi, G.~Anil, A.~Eck, P.~Doshi, and L.-K. Soh, ``Decision-theoretic planning with communication in open multiagent systems,'' in \emph{Uncertainty in Artificial Intelligence}.\hskip 1em plus 0.5em minus 0.4em\relax PMLR, 2022, pp. 938--948.

\bibitem{chandrasekaran2016individual}
M.~Chandrasekaran, A.~Eck, P.~Doshi, and L.~Soh, ``Individual planning in open and typed agent systems,'' in \emph{Proceedings of the Thirty-Second Conference on Uncertainty in Artificial Intelligence}, 2016, pp. 82--91.

\bibitem{eck2020scalable}
A.~Eck, M.~Shah, P.~Doshi, and L.-K. Soh, ``Scalable decision-theoretic planning in open and typed multiagent systems,'' in \emph{Proceedings of the AAAI Conference on Artificial Intelligence}, vol.~34, no.~05, 2020.

\bibitem{Goldman03}
C.~V. Goldman and S.~Zilberstein, ``Optimizing information exchange in cooperative multi-agent systems,'' in \emph{Proceedings of the Second International Joint Conference on Autonomous Agents and Multiagent Systems}, ser. AAMAS '03.\hskip 1em plus 0.5em minus 0.4em\relax New York, NY, USA: ACM, 2003.

\bibitem{melo2011decentralized}
F.~S. Melo and M.~Veloso, ``Decentralized mdps with sparse interactions,'' \emph{Artificial Intelligence}, vol. 175, no.~11, pp. 1757--1789, 2011.

\bibitem{fu2018learning}
J.~Fu, K.~Luo, and S.~Levine, ``Learning robust rewards with adverserial inverse reinforcement learning,'' in \emph{International Conference on Learning Representations}.\hskip 1em plus 0.5em minus 0.4em\relax unknown, 2018, pp. 1--10.

\bibitem{goldman2004decentralized}
C.~V. Goldman and S.~Zilberstein, ``Decentralized control of cooperative systems: Categorization and complexity analysis,'' \emph{Journal of artificial intelligence research}, vol.~22, pp. 143--174, 2004.

\bibitem{schulman2017proximal}
J.~Schulman, F.~Wolski, P.~Dhariwal, A.~Radford, and O.~Klimov, ``Proximal policy optimization algorithms,'' \emph{arXiv preprint arXiv:1707.06347}, 2017.

\bibitem{mcgugan2007beginning}
W.~McGugan, \emph{Beginning game development with Python and Pygame: from novice to professional}.\hskip 1em plus 0.5em minus 0.4em\relax Apress, 2007.

\bibitem{magym}
A.~Koul, ``ma-gym: Collection of multi-agent environments based on openai gym.'' \url{https://github.com/koulanurag/ma-gym}, 2019.

\bibitem{zhang2020}
F.~Zhang, V.~Bazarevsky, A.~Vakunov, A.~Tkachenka, G.~Sung, C.-L. Chang, and M.~Grundmann, ``Mediapipe hands: On-device real-time hand tracking,'' \emph{arXiv preprint arXiv:2006.10214}, 2020.

\bibitem{bartneck2008godspeed}
C.~Bartneck, D.~Kuli{\'c}, E.~Croft, and S.~Zoghbi, ``Godspeed questionnaire series,'' \emph{International Journal of Social Robotics}, 2008.

\bibitem{nikolaidis2012human}
S.~Nikolaidis and J.~Shah, ``Human-robot teaming using shared mental models,'' \emph{ACM/IEEE HRI}, 2012.

\bibitem{chen2020trust}
M.~Chen, S.~Nikolaidis, H.~Soh, D.~Hsu, and S.~Srinivasa, ``Trust-aware decision making for human-robot collaboration: Model learning and planning,'' \emph{ACM Transactions on Human-Robot Interaction (THRI)}, vol.~9, no.~2, pp. 1--23, 2020.

\bibitem{giacomuzzo2024decaf}
G.~Giacomuzzo, M.~Terreran, S.~Jain, and D.~Romeres, ``Decaf: a discrete-event based collaborative human-robot framework for furniture assembly,'' in \emph{IEEE/RSJ International Conference on Intelligent Robots and Systems (IROS)}, 2024.

\bibitem{nikolaidis2017game}
S.~Nikolaidis, S.~Nath, A.~D. Procaccia, and S.~Srinivasa, ``Game-theoretic modeling of human adaptation in human-robot collaboration,'' in \emph{Proceedings of the 2017 ACM/IEEE international conference on human-robot interaction}, 2017, pp. 323--331.

\bibitem{seo2022semi}
S.~Seo and V.~V. Unhelkar, ``Semi-supervised imitation learning of team policies from suboptimal demonstrations,'' \emph{arXiv preprint arXiv:2205.02959}, 2022.

\bibitem{wang2022co}
C.~Wang, C.~P{\'e}rez-D’Arpino, D.~Xu, L.~Fei-Fei, K.~Liu, and S.~Savarese, ``Co-gail: Learning diverse strategies for human-robot collaboration,'' in \emph{Conference on Robot Learning}.\hskip 1em plus 0.5em minus 0.4em\relax PMLR, 2022.

\bibitem{suresh2022marginal}
P.~S. Suresh and P.~Doshi, ``Marginal map estimation for inverse rl under occlusion with observer noise,'' in \emph{The 38th Conference on Uncertainty in Artificial Intelligence}, 2022.

\bibitem{chen2015considering}
B.~Chen, X.~Chen, A.~Timsina, and L.-K. Soh, ``Considering agent and task openness in ad hoc team formation.'' in \emph{AAMAS}, 2015.

\bibitem{hoang2013interactive}
T.~N. Hoang and K.~H. Low, ``Interactive pomdp lite: Towards practical planning to predict and exploit intentions for interacting with self-interested agents,'' \emph{arXiv preprint arXiv:1304.5159}, 2013.

\bibitem{cohen2017open}
J.~Cohen, J.-S. Dibangoye, and A.-I. Mouaddib, ``Open decentralized pomdps,'' in \emph{IEEE 29th International Conference on Tools with Artificial Intelligence (ICTAI)}, 2017.

\bibitem{rahman2023general}
A.~Rahman, I.~Carlucho, N.~H{\"o}pner, and S.~V. Albrecht, ``A general learning framework for open ad hoc teamwork using graph-based policy learning,'' \emph{Journal of Machine Learning Research}, vol.~24, 2023.

\end{thebibliography}

\end{document}